%% file: eacl2023.tex
\pdfoutput=1

\documentclass[11pt]{article}

\usepackage[]{EACL2023}

\usepackage{times}
\usepackage{latexsym}

\usepackage[T1]{fontenc}

\usepackage[utf8]{inputenc}

\usepackage{microtype}

\usepackage{inconsolata}

\usepackage{amsmath}
\usepackage{algorithm,algpseudocode}
\usepackage{mathtools}
\DeclareMathOperator*{\argmax}{argmax}

\usepackage{amssymb}
\DeclareMathSymbol{\shortminus}{\mathbin}{AMSa}{"39}

\usepackage{booktabs}
\usepackage{pifont}
\newcommand{\cmark}{\ding{51}}%
\newcommand{\xmark}{\ding{55}}%

\usepackage{cancel}
\newcommand{\Sref}[1]{\S\ref{#1}}
\newcommand{\Fref}[1]{Figure~\ref{#1}}
\newcommand{\Tref}[1]{Table~\ref{#1}}

\usepackage{multirow}

\newcommand\aspace{\hspace{1.2em}}

\title{CTC Alignments Improve Autoregressive Translation}

\author{
Brian Yan*$^1$ \aspace Siddharth Dalmia*$^1$ \aspace Yosuke Higuchi$^2$ \\ \textbf{Graham Neubig}$^1$ \aspace \textbf{Florian Metze}$^1$ \aspace \textbf{Alan W Black}$^1$ \aspace \textbf{Shinji Watanabe}$^{1,3}$ \\
$^1$Language Technologies Institute, Carnegie Mellon University, USA \\
$^2$Department of Communications and Computer Engineering, Waseda University, Japan \\
$^3$Human Language Technology Center of Excellence, Johns Hopkins University, USA ~~\\ 
\texttt{\{byan, sdalmia\}@cs.cmu.edu}
}

\begin{document}
\maketitle
\begin{abstract}
Connectionist Temporal Classification (CTC) is a widely used approach for automatic speech recognition (ASR) that performs conditionally independent monotonic alignment.
However for translation, CTC exhibits clear limitations due to the contextual and non-monotonic nature of the task and thus lags behind attentional decoder approaches in terms of translation quality. 
In this work, we argue that CTC \textit{does} in fact make sense for translation if applied in a joint CTC/attention framework wherein CTC's core properties can counteract several key weaknesses of pure-attention models during training and decoding.
To validate this conjecture, we modify the Hybrid CTC/Attention model originally proposed for ASR to support text-to-text translation (MT) and speech-to-text translation (ST).
Our proposed joint CTC/attention models outperform pure-attention baselines across six benchmark translation tasks.
\end{abstract}

\section{Introduction}

Automatic speech recognition (ASR), machine translation (MT), and speech translation (ST) have conspicuous differences but are all closely related sequence-to-sequence problems.
Researchers from these respective fields have long recognized the opportunity for cross-pollinating ideas \cite{he2011speech}, starting from the coupling of statistical ASR \cite{huang2014historical} and MT \cite{al1999statistical} which gave rise to early approaches for ST \cite{waibel1996interactive, ney1999}.
Notably in the end-to-end era, attentional encoder-decoder approaches emerged in both MT \cite{bahdanau_attn} and ASR \cite{chan2015listen} and have since risen to great prominence in both fields.

During this same period, there has been another prominent end-to-end approach in ASR: Connectionist Temporal Classification (CTC) \cite{graves2006connectionist}.
Unlike the highly flexible attention mechanism which can handle ASR, MT, and ST alike, CTC models sequence transduction as a monotonic alignment of inputs to outputs and thus fits more naturally with ASR than it does with translation.
Still, many interested in non-autoregressive translation have applied CTC to MT \cite{libovicky-helcl-2018-end} and ST \cite{chuang2021investigating} 
and promising techniques have emerged which have shrunk the gap between autoregressive approaches \cite{saharia-etal-2020-non, gu2021fully, inaguma2021non, huang2022non}.
These recent developments suggest that the latent alignment ability of CTC is a promising direction for translation --
this leads us to question: \textit{can CTC alignments improve autoregressive translation?}
In particular, we are interested in frameworks which leverage the strength of CTC while minimizing its several harmful incompatibilities (see \Sref{sec:ctc_limitations}) with translation tasks.

Inspired by the success of Hybrid CTC/Attention in ASR \cite{watanabe2017hybrid}, we investigate jointly modeling CTC with an autoregressive attentional encoder-decoder for translation. 
Our conjecture is that the monotonic alignment and conditional independence of CTC, which weaken purely CTC-based translation, counteract particular weaknesses of attentional models in joint CTC/attention frameworks.
In this work, we seek to investigate \textit{how} each CTC property interacts with corresponding properties of the attentional counterpart during joint training and decoding.
We design a joint CTC/attention architecture for translation (\Sref{sec:model}) and then examine the positive interactions which ultimately result in improved translation quality compared to pure-attention baselines, as demonstrated on the IWSLT \cite{cettolo2012wit3}, MuST-C \cite{di-gangi-etal-2019-must}, and MTedX \cite{salesky2021multilingual} MT/ST corpora (\Sref{sec:results}).

\section{Background: Joint CTC/Attn for ASR}
\label{sec:background}

\label{sec:hybrid_asr}

\begin{table*}[t]
  \centering
    \include{tables/conjectures}
    \caption{Description of three reasons why joint CTC/attention modeling is powerful in ASR. In order to understand whether these positive interactions between properties of the \textcolor{blue}{CTC} and \textcolor[HTML]{b2182b}{attention} frameworks are applicable to MT/ST, we must address three corresponding concerns, L1-3, about the applicability of CTC to translation (\Sref{sec:background}).}
    \label{tab:conjectures}
\end{table*}

Both the CTC \cite{graves2006connectionist} and attentional encoder-decoder \cite{bahdanau_attn} frameworks seek to model the Bayesian decision seeking the output, $\hat{Y}$, from all possible sequences, $\mathcal{V}^\text{tgt}*$, by selecting the sequence which maximizes the posterior likelihood $P(Y|X)$, where $X = \{ \mathbf{x}_t \in \mathcal{S}^\text{src} | t = 1, ..., T \}$ and $Y = \{ y_l \in \mathcal{V}^\text{tgt} | l = 1, ..., L \}$. As shown in the first two columns of \Tref{tab:conjectures}, CTC and attention offer different formulations of the posterior likelihood, $P_\text{CTC}(\cdot)$ and $P_\text{Attn}(\cdot)$ respectively.

\textit{What are the critical differences between the CTC and attention frameworks?}
First of all, the attention mechanism is a flexible input-to-output mapping function which allows a decoder to perform \textcolor[HTML]{b2182b}{\textbf{soft alignment}} of an output unit $y_l$ to multiple input units $\mathbf{x}_{[...]}$ without restriction.
One downside of this flexibility is a risk of destabilized optimization \cite{kim2017joint}.
CTC on the other hand marginalizes the likelihoods of all possible input to alignment sequence, $Z = \{ z_t \in \mathcal{V}^\text{tgt} \cup \{\varnothing \} | t=1\ldots T \}$, mappings via \textcolor{blue}{\textbf{hard alignment}} where each output unit $z_t$ maps to a single input unit $\mathbf{x}_t$ in a strictly monotonic pattern.\footnote{$\varnothing$ is a "blank" denoting null emission and $Z$ maps deterministically to $Y$ by removing null and repeated emissions.}

Secondly, the attentional decoder models each output unit $y_1$ with \textcolor[HTML]{b2182b}{\textbf{conditional dependence}} on not only the input $X$, but also the previous output units $y_1:l-1$.
In order to efficiently compute the marginalized likelihoods of all possible $Z \in \mathcal{Z}(Y,T)$ via dynamic programming, CTC makes a \textcolor{blue}{\textbf{conditional independence}} assumption that each $z_t$ does not depend on $z_{1:t}$ if already conditioned on $X$ - this is a strong assumption. 
On the plus, since CTC does not model any causality between output units its is not plagued by the same label and exposure biases that exist in attentional decoders due to local normalization of causal likelihoods \cite{bottou-91a, ranzato2016sequence, awnilabelbias}. 

Finally, the attentional decoder is an \textcolor[HTML]{b2182b}{\textbf{autoregressive generator}} that decodes the output until a stop token, $\text{<eos>}$.
Comparing likelihoods for sequences of different lengths requires a heuristic brevity penalty. Furthermore label bias with respect to the stop token manifests as a length problem where likelihoods degenerate for unexpectedly long outputs \cite{murray2018correcting}.
In comparison, CTC is an \textcolor{blue}{\textbf{input-synchronous emitter}} that consumes an input unit in order to produce an output unit.
Therefore, CTC cannot produce an output longer than the input representation which feeds the final posterior output layer but this also means that CTC does not require any end detection.  

As previously shown by \cite{kim2017joint, watanabe2017hybrid}, jointly modeling CTC and an attentional decoder is highly effective in ASR. 
The foundation of this architecture is a shared encoder, $\textsc{Enc}$, which feeds into both CTC, $P_\text{CTC}(\cdot)$, and attentional decoder, $P_\text{Attn}(\cdot)$, posteriors:
\begin{align}
    \mathbf{h} &= \operatorname{Enc}(X) \\
    P_\text{CTC}(z_t | X) &= \operatorname{CTC}(\mathbf{h}_t) \\
    P_\text{Attn}(y_l | X, y_{1:l-1}) &= \operatorname{Dec}(\mathbf{h}, y_{1:l-1})
\end{align}
where $\textsc{CTC}(\cdot)$ denotes a projection to the CTC output vocabulary, $\mathcal{V}^\text{tgt} \cup \{\varnothing \}$ followed by softmax, and $\textsc{Dec}(\cdot)$ denotes autoregressive decoder layers followed by a projection to the decoder output vocabulary, $\mathcal{V}^\text{tgt} \cup \{\text{<eos>} \}$, and softmax.
The joint network is optimized via a multi-tasked objective, $\mathcal{L}^\text{ASR} = \mathcal{L}^\text{ASR}_\text{CTC} + \lambda \mathcal{L}^\text{ASR}_\text{Attn}$,
where $\lambda$ interpolates the CTC loss and the cross-entropy loss of the decoder.

Joint decoding is typically performed with a one-pass beam search where CTC plays a secondary role as a joint scorer while attention leads the major hypothesis expansion and end detection functions in the algorithm \cite{watanabe2017hybrid, tsunoo2021}.
However, CTC is capable of taking over the lead role if called upon (e.g. for streaming applications) \cite{moritz2019}.

\section{Potential CTC Limitations in MT/ST}
\label{sec:ctc_limitations}
\textit{Why exactly does this joint CTC/attention framework perform so well in ASR?}
As summarized in column 3 of \Tref{tab:conjectures}, we are particularly interested in three reasons which arise from the combination of the hard vs. soft alignment, conditional independence vs. dependence, and input-synchronous emission vs. autoregressive generation properties of CTC and attention respectively.
These dynamics have become well understood in ASR, owing to the popularity of the joint framework \cite{watanabe2018espnet} amongst ASR practitioners.

\textit{So can CTC and attention also complement each other when applied jointly to translation?}\footnote{This particular question has not been addressed in literature. For an account of related works, please see \Sref{sec:prior_works}.}
ASR, MT, and ST can all be generalized as sequence transduction tasks following the Bayesian formulation. 
Plus attentional decoders have are a predominant technical solution to each of these tasks.
However, the CTC framework appears to have several limitations specific to MT/ST that are not present in ASR; this seemingly diminishes the promise of the joint CTC/attention framework for translation.
In this work, we seek to address the following three concerns about MT/ST CTC which appear to inhibit the CTC/attention framework (per \Tref{tab:conjectures}).

\paragraph{L1} \textit{Can CTC encoders perform sophisticated input-to-output mappings required for translation?}

Unlike ASR, translation entails non-monotonic mappings due to variable word-ordering across languages.
Additionally, inputs may be shorter than outputs as mappings are not necessarily one-to-one.
Furthermore, the mapping task for ST is actually compositional where logically a speech signal first maps to a source language transcription before being mapped to the ultimate translation.
All of these complications appear to directly contradict the \textcolor{blue}{\textbf{hard alignment}} of CTC.
If CTC cannot produce stable encoder representations for MT/ST, then during joint training attention does not receive the optimization benefit as in ASR (per row 2 of \Tref{tab:conjectures}).
Fortunately, prior works suggest that these challenges are not insurmountable.
\citet{chuang2021investigating} showed that self-attentional encoders can perform latently model variable word-orders for ST, \citet{libovicky-helcl-2018-end, dalmia2022legonn} proposed up-sampling encoders that produce expanded input representations for MT, and \citet{sanabria2018hierarchical, higuchi2022hierarchical} proposed hierarchical encoders that can compose multiple output resolutions for ASR.
In \Sref{sec:encoding}, we incorporate these techniques into a unified solution which achieves hierarchical encoding for translation.

\paragraph{L2} \textit{Does CTC-based translation quality lag too far behind attention-based to be useful?}

CTC-based ASR has recently shown competitive performance due in large part to improved neural architectures \cite{gulati2020conformer} and self-supervised learning \cite{baevski2020wav2vec, hsu2021hubert}, but the gap between CTC and attention for translation appears to be greater \cite{gu2021fully}.
Perhaps the \textcolor{blue}{\textbf{conditional independence}} of CTC inhibits the quality to such a degree in MT/ST where these likelihoods cannot ease the label/exposure biases of the attentional decoder as they do in ASR (per column 3 of \Tref{tab:conjectures}). 
The relative weakness of non-autoregressive translation approaches has been well-studied.
Knowledge distillation \cite{kim2016sequence, zhou2019understanding} and iterative methods \cite{qian2021glancing, chan2020imputer, huang2022non} all attempt to bridge the gap between non-autoregressive models and their autoregressive counterparts.
In \Sref{sec:results}, we address this concern empirically: even CTC models with 28\% relative BLEU reduction compared to attention yield improvements when CTC and attention are jointly decoded.

\paragraph{L3} \textit{Is the alignment information produced by CTC-based translation models reasonable?}

In ASR, CTC alignments are reliable enough to segment audio data by force aligning inputs to a target transcription outputs \cite{kurzinger2020ctc} and exhibits minimal drift compared to hidden Markov models \cite{sak2015learning}.
However, CTC alignments are not as well studied in translation.
It is an open question of whether or not the \textcolor{blue}{\textbf{input-synchronous emission}} of CTC for translation has sufficient alignment quality to support the end detection responsibility during joint decoding as it does in ASR (per row 4 of \Tref{tab:conjectures}).
Ideally, the CTC alignments are strong enough such that CTC can lead joint decoding by proposing candidates for hypothesis expansion in each beam step until all input units are consumed (at which point the end is detected), as in an input-synchronous beam search.
More conservatively, the CTC alignments may be too unreliable to take lead but could still guide the attentional decoder's end detection by penalizing incorrect lengths via joint scoring, as in an output-synchronous beam search.
In \Sref{sec:decoding}, we lay out comparable forms for input and output-synchronous beam search which allows us to examine the impact on translation quality depending on whether CTC is explicitly responsible for or only implicitly contributing to end detection.

\section{Joint CTC/Attention for Translation}
\label{sec:model}

\subsection{Hierarchical CTC Encoding}
\label{sec:encoding}

\begin{figure}
    \centering
    \includegraphics[width=.95\linewidth]{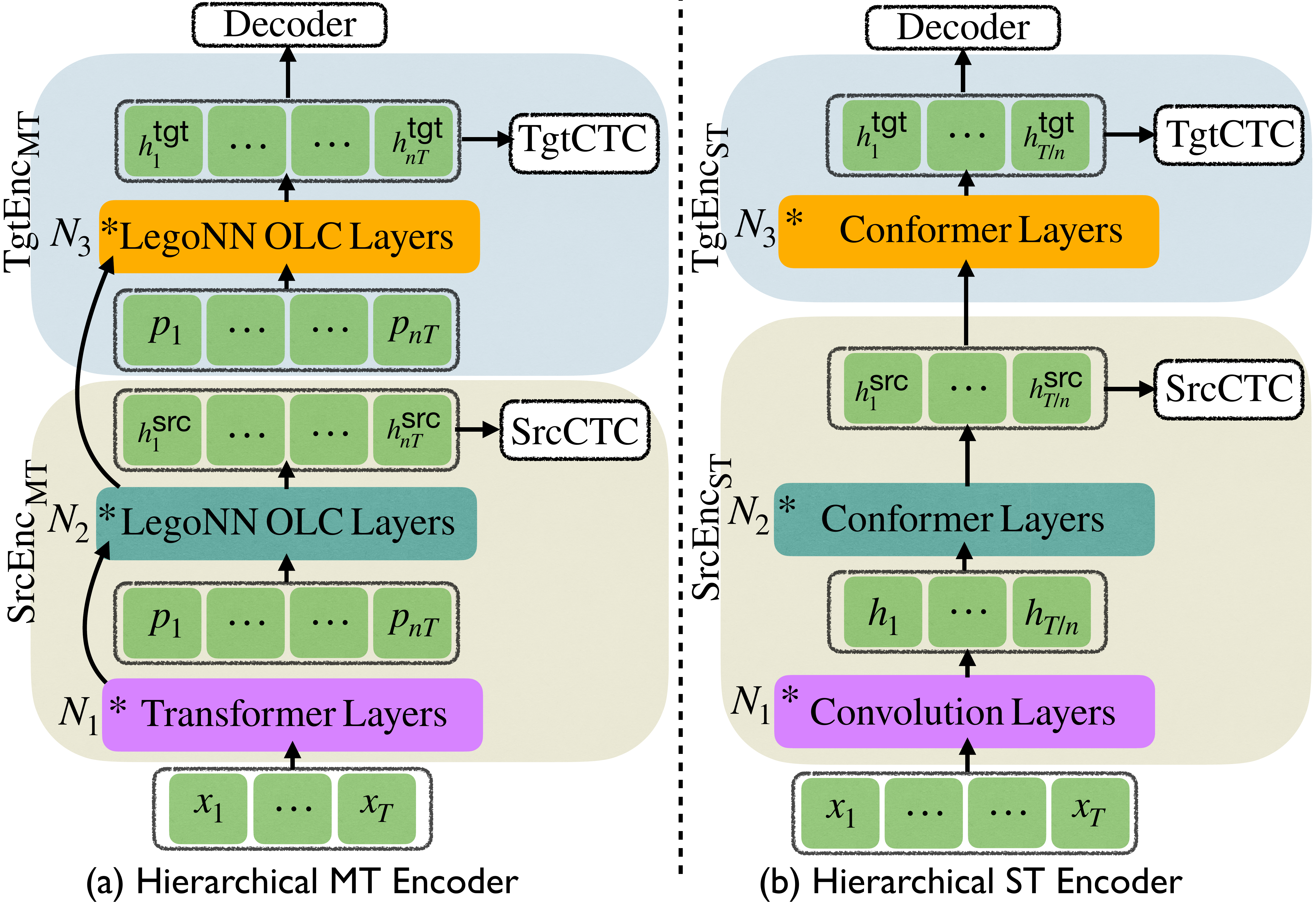}
    \caption{Hierarchical MT/ST encoders where representations are first up/down-sampled by $\textsc{SrcEnc}_\text{MT/ST}$ and then re-ordered by $\textsc{TgtEnc}_\text{MT/ST}$.}
    \label{fig:model}
\end{figure}

Per L1 described in \Sref{sec:ctc_limitations}, we seek to build a CTC encoder for translation which handles sophisticated input-to-output mappings.
We therefore propose to use a hierachical CTC encoding scheme\footnote{Hierarchical CTC encoding is a flexible technique which has been applied for various multi-objective scenarios in prior ASR works \cite{sanabria2018hierarchical, higuchi2022hierarchical}.} which first aligns inputs to length-adjusted \textit{source}-oriented encodings before aligning to re-ordered \textit{target}-oriented encodings, as shown in \Fref{fig:model}.
We decompose the encoding process into two functions: length-adjustment and re-ordering.

\paragraph{Length-adjustment}
For MT, we up-sample the lengths of the source-oriented encodings in order to output sequences longer than the input.
For ST, we down-sample the lengths of the source-oriented encodings to coerce a discrete textual representation of the real-valued speech input.
We enforce source-orientations using CTC criteria that seek to align intermediate layer encoder representations towards source text sequences.

\paragraph{Re-ordering}
We then obtain target-oriented encodings with hierarchical encoder layers, where re-ordering is enforced using CTC criteria that seek to align final layer encoder representations towards target text sequences.
Critically, the underlying neural network architecture must be able to model latent re-ordering as the CTC criterion itself will only consider monotonic alignments of the final encoder representation to the target.

Our proposed MT/ST hierarchical encoders consist of the following components:
\begin{align}
    \mathbf{h}^{\textsc{Src}} &= \textsc{SrcEnc}_\text{MT/ST}(X) \\
    P_\textsc{CTC}(z^{\textsc{Src}}_t | X) &= \textsc{SrcCTC}_\text{MT/ST}(\mathbf{h}^{\textsc{Src}}_t) \\
    \mathbf{h}^{\textsc{Tgt}} &= \textsc{TgtEnc}_\text{MT/ST}(\mathbf{h}^{\textsc{Src}}) \\
    P_\textsc{CTC}(z^{\textsc{Tgt}}_t | X) &= \textsc{TgtCTC}_\text{MT/ST}(\mathbf{h}^{\textsc{Tgt}}_t)
\end{align}
where $\textsc{SrcEnc}_\text{MT}(\cdot)$ is realized by $N_1$ Transformer \cite{vaswani2017attention} layers followed by $N_2$ up-sampling LegoNN Output Length Controller (OLC) layers \cite{dalmia2022legonn}, while $\textsc{TgtEnc}_\text{MT}(\cdot)$ is realized by $N_3$ non-up-sampling LegoNN OLC layers.
We chose LegoNN based on its previously demonstrated effectiveness for up-sampling textual representations and its ability to perform latent re-ordering via self-attention.
$\textsc{SrcEnc}_\text{ST}(\cdot)$ is realized by $N_1$ convolutional blocks for downsampling \cite{dong2018speech} followed by $N_2$ Conformer \cite{gulati2020conformer}, while $\textsc{TgtEnc}_\text{ST}(\cdot)$ is realized by $N_3$ Conformer layers.
We chose Conformer based on its previously demonstrated effectiveness for modeling local and global dependencies in speech signals and its ability to perform latent re-ordering via self-attention.

The hierarchical encoders are jointly optimized with an attentional decoder using a multi-tasked objective, $\mathcal{L} = \mathcal{L}_\textsc{SrcCTC} + \lambda_1 \mathcal{L}_\textsc{TgtCTC} + \lambda_2 \mathcal{L}_\textsc{Attn}$,
where $\lambda$'s interpolate source-oriented CTC, target-oriented CTC, and decoder cross-entropy losses.

\begin{figure*}
  \centering
  \includegraphics[width=\textwidth]{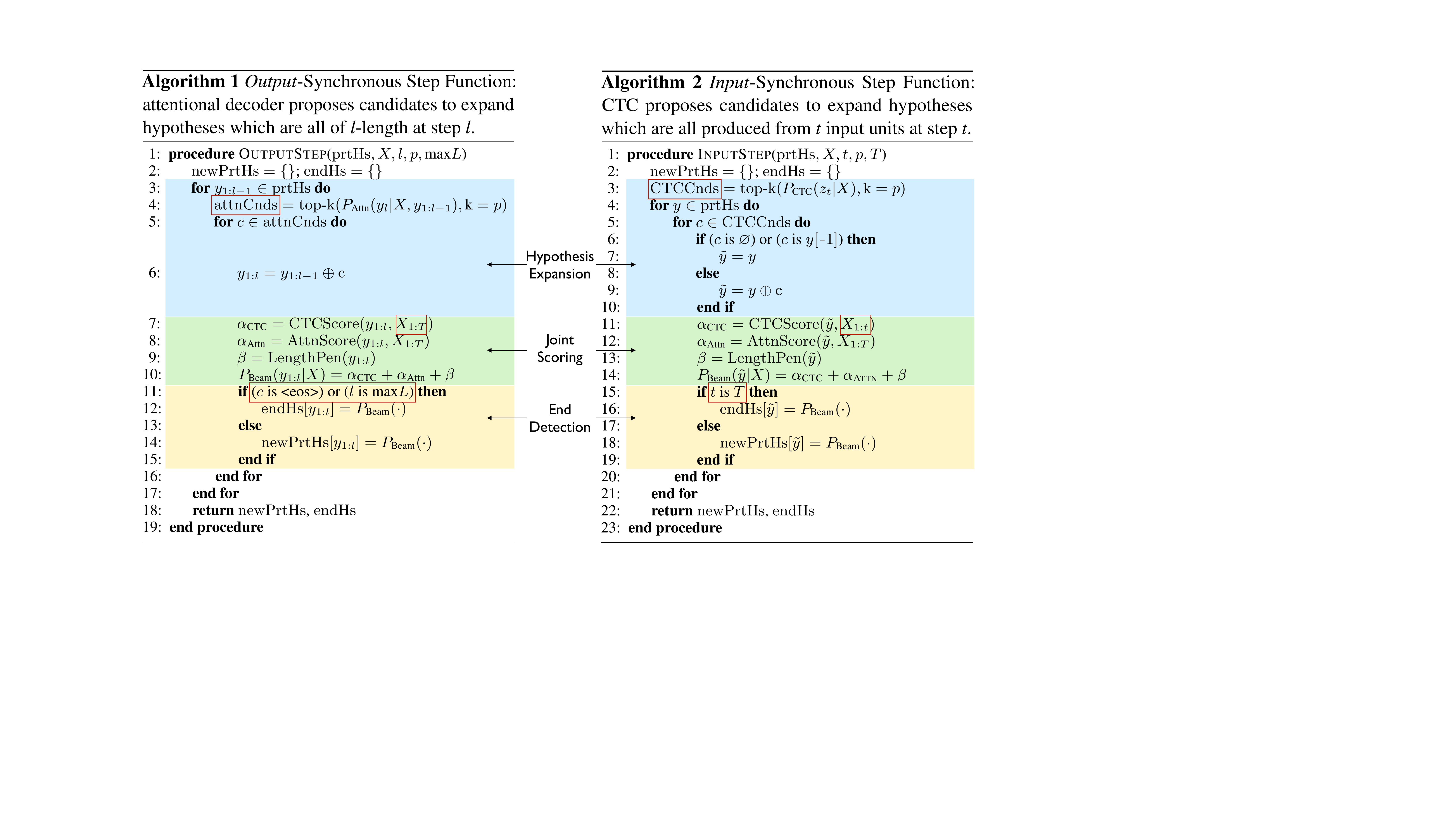}
\end{figure*}

\subsection{Input/Output-Synchronous Decoding}
\label{sec:decoding}

Per L2 and L3 described in \Sref{sec:ctc_limitations}, we seek to design a joint decoding algorithm with input and output-synchronous variants of one-pass beam search which differ only in whether CTC or attention takes the leading role.
As shown in Algorithms 1 and 2, we propose to align the input and output beam-step functions along three common functions: hypothesis expansion, joint scoring, and end detection.

\paragraph{Output-Synchrony} Consider first that attention is in the leading role, which means that we are working with an output-synchronous beam search.
Note that this is the algorithm originally proposed by \citet{watanabe2017hybrid}.
$\textsc{OutputStep}$ performs \textit{hypothesis expansion} by computing the attentional decoder's output posterior at label step $l$, $P_{\text{Attn}}( y_l | X,y_{1:l-1})$ for each partial hypothesis, $y_{1:l-1}$.
A pre-beam size, $p$, is then used to select the top candidate output units \cite{seki2019vectorized}, $\operatorname{attnCnds}$, which are used to expand the partial hypotheses via concatenation, denoted by $\oplus$.
In the \textit{joint scoring} block, the attentional decoder likelihood, $\operatorname{AttnScore}(\cdot)$, and length penalty/reward, $\operatorname{LengthPen}(\cdot)$ yield the estimated joint likelihood $P_\text{Beam}$.
Finally in \textit{end detection}, $\textsc{OutputStep}$ must check for the stop token, $\text{<eos>}$, which may be proposed by $\operatorname{attnCnds}$.

\paragraph{Input-Synchrony}
Now let us consider the differences when CTC is in the leading role.
$\textsc{InputStep}$ performs \textit{hypothesis expansion} by computing CTC's alignment posterior at time step $t$, $P_{\text{CTC}}(z_t| X)$.
Unlike in output-synchrony, here each hypothesis expansion also consumes one step of the input.
The same pre-beam size, $p$, is used to select top candidate alignment units, $\operatorname{CTCCnds}$, but partial hypotheses are only expanded for non-blank and non-repeat candidates.
The \textit{joint scoring} block is identical to output-synchrony except for one difference: CTC likelihood, $\operatorname{CTCScore}(\cdot)$, is applied over the full input, $X_{1:T}$, in $\textsc{OutputStep}$ and over the partial input, $X_{1:t}$, in $\textsc{InputStep}$.
This difference engenders a speed vs. accuracy trade-off, which we discuss in D2 of \Sref{sec:discussion}.
Finally, \textit{end detection} simply occurs when all input units have been consumed ($t=T$).
Therefore, $\textsc{InputStep}$ does not require checking for the stop token as all hypotheses at time $T$ are ended. 

We propose this particular form of input-synchronous beam search in order to exactly mirror the functions of its the output-synchronous counterpart; without this mirrored formulation, we cannot attribute differences in decodings to the swapped roles of CTC and attention.
For instance, Triggered Attention \cite{moritz2019triggered} is another form of input-synchronous CTC/attention beam search but since it is purpose-fit for streaming it is not trivial to draw parallels to an output-synchronous variant.
This method contains look-ahead and re-triggering strategies which are tightly coupled to input consumption and further pruning may be done using de-synchronized joint scores.

\begin{table*}[t]
  \centering
    \include{tables/main}
    \caption{Test set performances, as measured by BLEU ($\uparrow$), of our proposed joint CTC/Attention models compared to pure-attention baselines. Joint CTC/Attention models are always jointly trained, but can be either jointly decoded using input/output synchronony or decoded using only their CTC or attention branches. For IWSLT14, we mention (\textit{tokenized BLEU}) for comparison with prior works: ~$^\dag$\citet{raunak2020long} and $^\ddag$\citet{inaguma2020espnet}.
    $^\diamondsuit$Prior MTedX works show only All-All or pair-wise settings.
    }
    \label{tab:main}
\end{table*}

\section{Experimental Setup}
\paragraph{Data} We examine the efficacy of our proposed approaches on two language pairs for each of the MT and ST tasks. For MT, we use German-to-English (De-En) and Spanish-to-English (Es-En) from IWSLT14~\cite{cettolo2012wit3}. For ST, we use English-to-German (En-De) and English-to-Japanese (En-Ja) from MuST-C-v2, reporting tst-COMMON results \cite{di-gangi-etal-2019-must}. We also examine the multilingual setting of 6 European languages to English (All-En) from MTedX \cite{salesky2021multilingual} for both tasks. Full dataset descriptions for reproducibility are in \Sref{sec:repro}.

\paragraph{Modeling} We compare our joint CTC/Attention models to purely attentional encoder-decoder baselines. All proposed and baseline models were tuned separately, using validation sets only, within the same hyperparameter search spaces for training and decoding to ensure fair comparison. 
All experiments were conducted using ESPnet-ST \cite{inaguma2020espnet}. Full descriptions of model sizes, hyperparameters, and pre-processing are in \Sref{sec:repro}.\footnote{We compare our baselines for MuST-C-v2 to the default recipes in ESPnet in \Tref{tab:main}. For back-compatibility with additional prior works using MuST-C-v1 En-De, see \Sref{sec:v1}.}

\paragraph{Evaluation:} Unless otherwise indicated, we measure performance with detokenized case-sensitive BLEU \cite{post-2018-call} on punctuated 1-references. Detailed on scoring tools are in \Sref{sec:repro}.

\section{Results and Analyses}
\label{sec:results}

In this section, we first present our main results on 6 benchmark MT and ST tasks.
We then present evidence that hierarchical encoding (\Sref{sec:encoding}) produces stable encoder representations that simplify the decoder's source attention (addressing L1 in \Sref{sec:ctc_limitations}).
Next we present evidence that joint decoding is beneficial despite the fact that CTC-only performance lags behind that of attention-only (addressing L2 in \Sref{sec:ctc_limitations}).
Finally, we present evidence that CTC's alignment information resolves attention's end-detection problem in both input and output synchronous joint decoding (\Sref{sec:decoding}) (addressing L3 in \Sref{sec:ctc_limitations}).

\subsection{Joint CTC/Attention Models Outperform CTC-only and Attention-only Baselines}
\label{sec:main_res}
As shown in \Tref{tab:main}, joint CTC/Attention outperforms pure-attention across 3 MT tasks (IWSLT14 De-En + Es-En and MtedX All-En) and 3 ST tasks (MuST-C-v2 En-De + En-Ja and MTedX All-En).
In the second horizontal partition, we show that joint training while only decoding with the attention branch outperforms pure-attention models without any joint training.
Note that \textit{CTC is consistently the weaker} of the two branches in jointly trained models.
In the last horizontal partition, we show that joint input/output-synchronous decodings yield further improvements, confirming that \textit{both joint training and decoding are beneficial}.

\subsection{Hierarchical Encoding Reduces Attention's Alignment Burden}
\begin{figure}
\centering
\includegraphics[width=.95\linewidth]{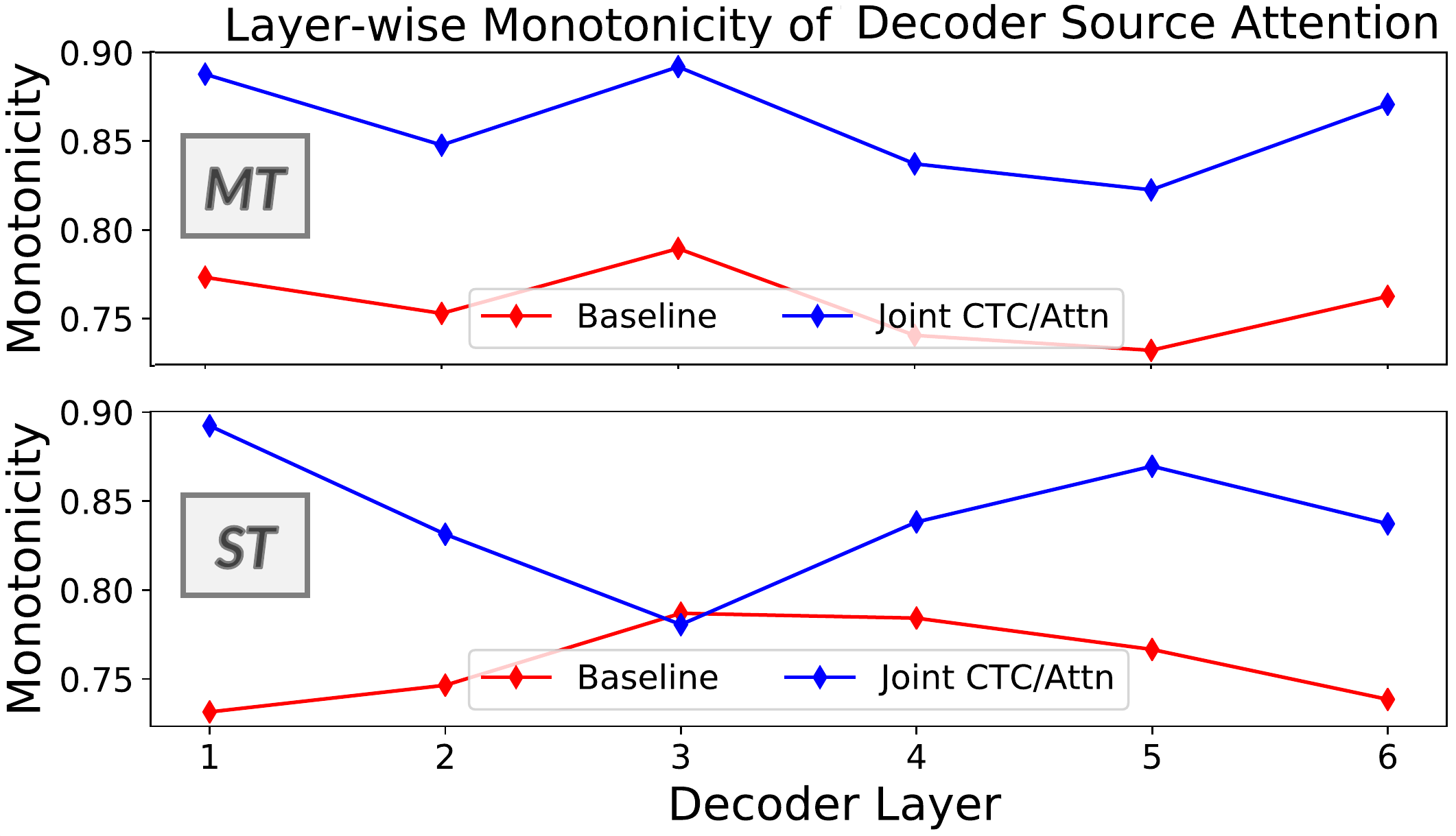}
\caption{Layer-wise monotonicity of the source-attention patterns produced by MT/ST decoders.}
\label{fig:mono}
\end{figure}

We examine the regularization effect that CTC joint training has on the attentional decoder, per L1 in \Sref{sec:ctc_limitations}, by first quantifying the monotonicity, $m$ of a $(L, T)$ shaped source attention pattern, $A$: 
\begin{equation*}
    m = \Big( \sum_{2 < l \le L} [\argmax_{t \in T} A_{l} \ge \argmax_{t \in T} A_{l-1} ] \Big) / L
\end{equation*}
where $[\cdot]$ denotes the Iverson bracket.
We compute $m$ over all examples in our validation sets for De-En MT and En-De ST and show the layer-wise averages over all examples and attention heads in \Fref{fig:mono}.
It can be seen that \textit{the decoder source attention patterns are more monotonic} when using jointly trained hierarchical encoders.

\begin{figure}
\centering
\includegraphics[width=0.9\linewidth]{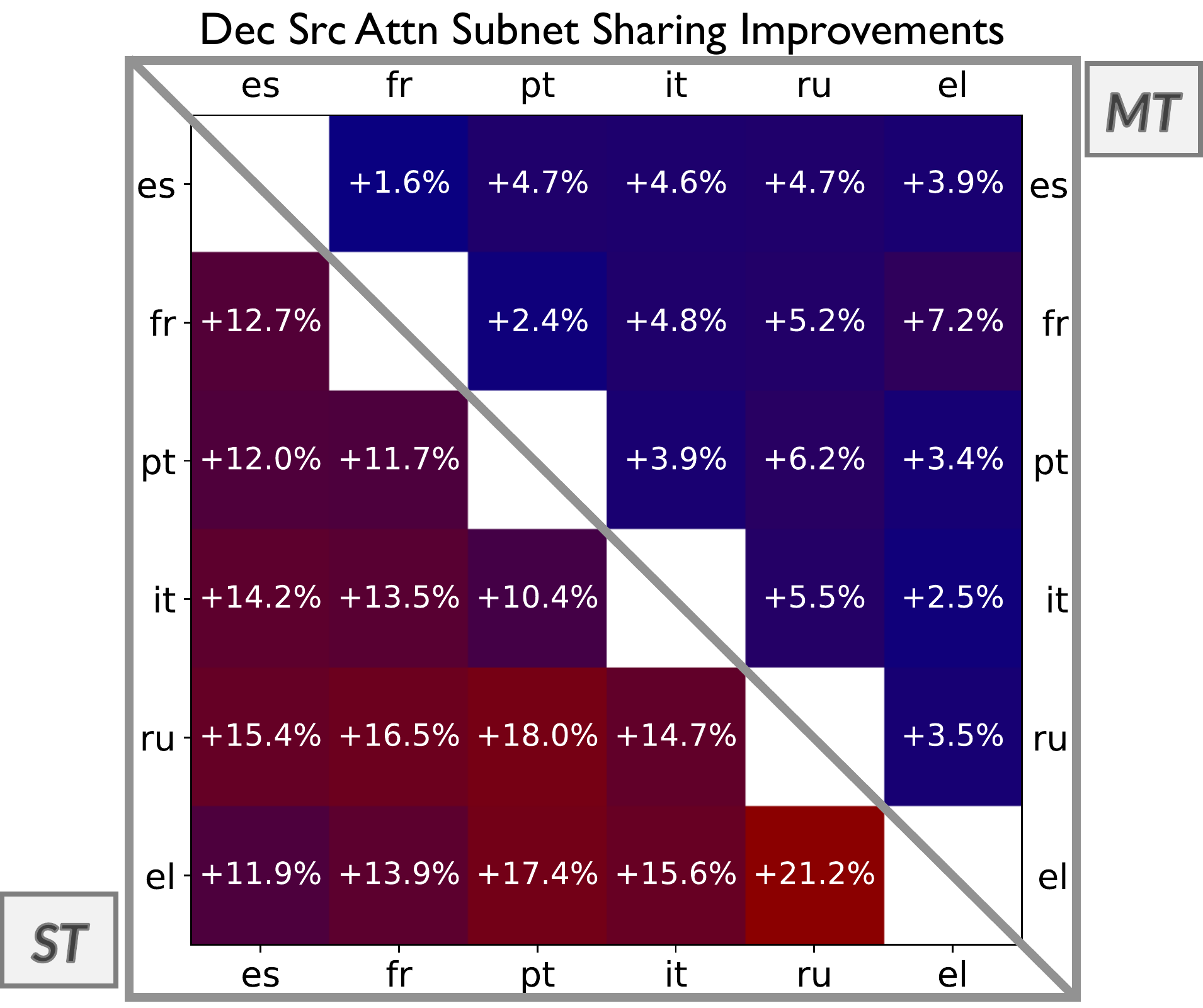}
\caption{Improvement of multilingual sharing in MT/ST decoder source attention parameters when using joint CTC/Attention vs. attention-only training, as measured by pair-wise $\Delta\%$ in sparse subnet overlap.}
\label{fig:subnet}
\end{figure}

We further examine the source attention parameters in our All-En models to understand the impact that the aforementioned regularization has on multilingual parameter sharing.
To do so, we extract sparse subnets for each language pair following the Lottery Ticket Sparse Fine-Tuning proposed by \citet{ansell2022composable} and compute the pair-wise sharing across the 6 source languages, as measured by the count of overlapping parameters between subnets.
In \Fref{fig:subnet}, we show the relative change ($\Delta\%$) in multilingual sharing when using hierachical encoding compared to the baseline.
The broad increases suggest that the target-orientation of our encoder \textit{reduced the decoder's burden of soft-aligning} target English outputs to source languages with varying word-orders, allowing for more efficient allocation of modeling capacity.

\textit{What are the respective contributions of \textsc{SrcCTC} and \textsc{TgtCTC}?}
\textsc{TgtCTC} holds elevated importance as joint decoding is not possible without it.
However, we'd like to understand how both components contribute to the benefits observed from joint training without joint decoding in \Sref{sec:main_res}.
In \Tref{tab:ablation}, we show ablate \textsc{SrcCTC} and \textsc{TgtCTC} in order to confirm that both contribute to performance gains.
Note that \textsc{SrcCTC} on its own appears to contribute more to MT than it does to ST, suggesting that the length adjustment stage is more critical in MT.

\subsection{Even Weak CTC Models Strengthen Joint CTC/Attention Models}
\label{sec:sync_res}

\begin{table}[t]
  \centering
    \include{tables/ablation}
    \caption{Ablation on the impacts of \textsc{SrcCTC} and \textsc{TgtCTC} CTC components of hierarachical encoding, as measured by performance on validation sets. Only attention is used in decoding for fair comparison.}
    \label{tab:ablation}
\end{table}

\begin{table}[t]
  \centering
    \include{tables/out_domain}
    \caption{In/out domain test performances of joint CTC/attention models with various decoding methods: decoding each branch separately, decoding one first and then re-scoring, or decoding both with synchrony.}
    \label{tab:out_domain}
\end{table}

We examine the generalization effect that augmenting autoregressive likelihoods with conditionally independent likelihoods has during inference, per L2 in \Sref{sec:ctc_limitations}, by evaluating De-En MT and En-De ST models on out-of-domain EuroParl test sets \cite{iranzo2020europarl}.
As shown in \Tref{tab:out_domain}, joint CTC/Attention models outperform pure-attention baselines across in-domain (In-D) and out-of-domain (Out-D) settings.
When decoding only the CTC branch of joint models (denoted as CTC I-Sync in the table) performance is significantly degraded compared to the attention branch of the same models (denoted as Attn O-Sync in the table). 
This gap appears slightly lessened on the out-of-domain setting where CTC's conditional independence may offer some robustness.
Nonetheless these weak CTC models still boost their stronger attention counterparts during joint decoding (both via input and output-synchrony), suggesting that \textit{ensembling of conditionally independent and dependent likelihoods is a powerful technique}.

Further, synchronous joint decoding methods outperform their two-pass re-scoring counterparts (discussed in D2 of \Sref{sec:discussion}), suggesting that \textit{joint selection of the hypothesis set is necessary} for easing the respective weaknesses of autoregressive and conditionally independent likelihood estimation.

\subsection{CTC's Alignment Information Resolves Attention's End-Detection Problem}
\begin{figure}
\centering
\includegraphics[width=\linewidth]{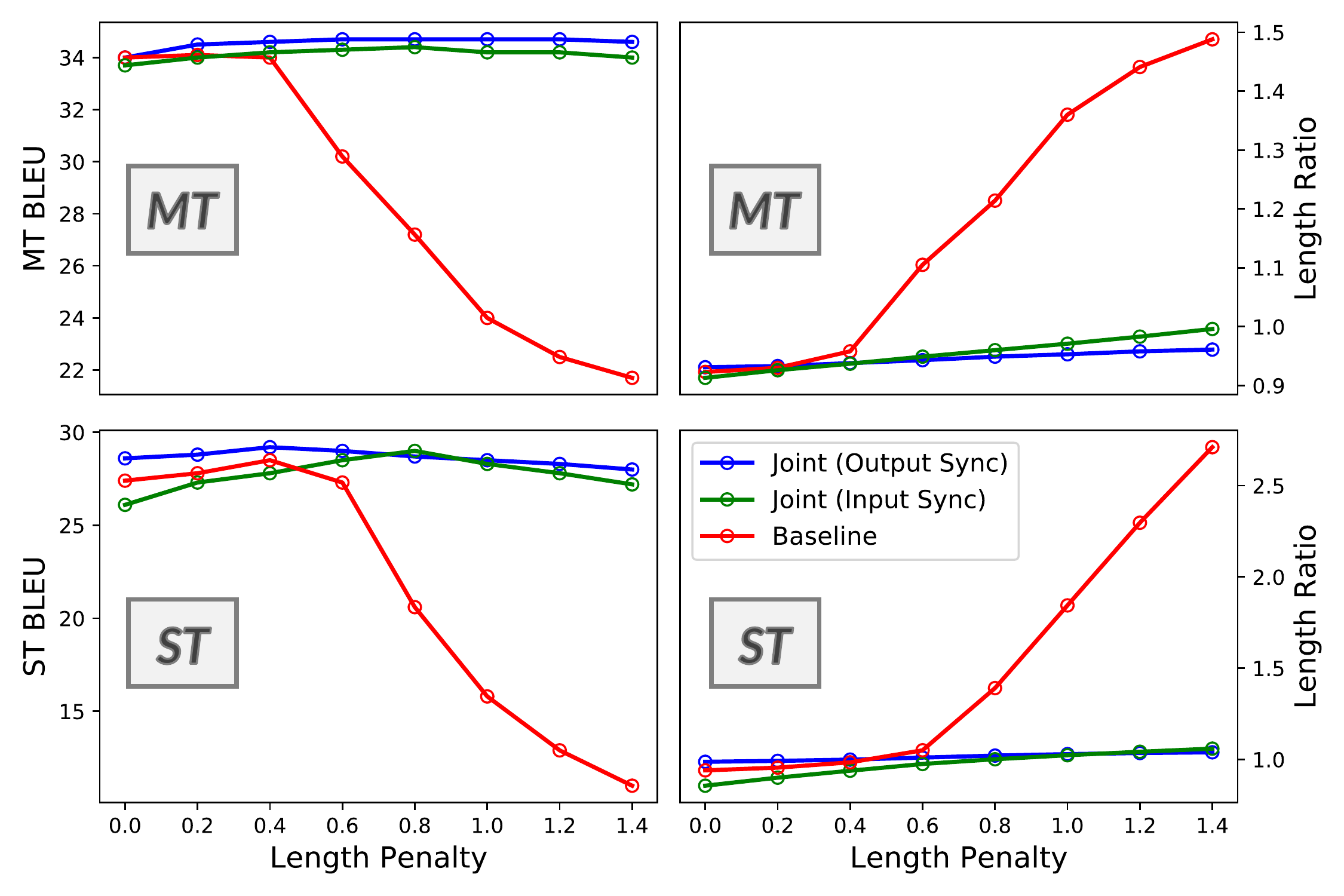}
\caption{Elasticity of BLEU and length ratios ($|\text{hyp}|/|\text{ref}|$) w.r.t length penalty in validation sets.}
\label{fig:pen}
\end{figure}

Finally, we examine the effect that CTC's alignment information has end detection during decoding, per L3 in \Sref{sec:ctc_limitations}.
In \Fref{fig:pen}, we observe the change in translation quality (as measured by BLEU) and output length (as measured by hypothesis-to-reference length ratio) when the length penalty term (denoted as $\operatorname{LengthPen(\cdot)}$ in Algorithms 1 and 2) is gradually increased, forcing decodings to produce longer outputs.
Pure-attention MT/ST baselines rapidly degenerate when forced to produce hypotheses that are longer than references as they struggle to detect the ends of hypotheses.
On the other hand, synchronous joint decoding produces gradually longer outputs regardless of whether CTC is in a leading role (input-synchrony) or a secondary role (output-synchrony),
demonstrating that \textit{CTC alignments eases the decoder's end-detection problem} by explicitly or implicitly ruling out hypotheses of incorrect lengths.

\section{Discussion: More on Joint Decoding}
\label{sec:discussion}

\begin{table}[t]
  \centering
    \include{tables/label_vs_time}
    \caption{Speed vs. accuracy for joint input/output-sync decoding of En-De ST val. set as a fxn. of beam size.}
    \label{tab:label_vs_time}
\end{table}

\paragraph{D1} \textit{Why do input vs. output-synchronous joint decodings yield slightly different results?}

By comparing the CTC likelihood estimation in $\textsc{InputStep}$ vs. $\textsc{OutputStep}$, it can be seen that there is a trade-off between speed vs. accuracy.
First, note that in $\textsc{OutputStep}$, $\operatorname{CTCScore}(y_{1:l}, X_{1:T})$, is a marginalization over the likelihoods of all possible alignments of the partial hypothesis $y_{1:l}$ to the full input $X_{1:T}$.
On the other hand, $\operatorname{CTCScore}(\tilde{y}, X_{1:t})$ in $\textsc{InputStep}$ is an estimation of the marginalized likelihoods of the partial hypothesis $y_{1:l}$ to the \textit{partial} input $X_{1:t}$.
Even at step $T$, these two $\operatorname{CTCScore}(\cdot)$'s are not equivalent. 
Since $\operatorname{CTCCnds}$ may include the blank token, $\textsc{InputStep}$ may prune partial hypotheses at a previous beam step which would have merged with $y_{1:l}$.
Therefore, $\operatorname{CTCScore}(\cdot)$ in \textit{input-synchrony is less accurate}.
However, \textit{input-synchrony requires fewer computations}.
Using dynamic programming, output-synchrony computes $\operatorname{CTCScore}(\cdot)$ for all partial hypothesis within a single beam step with $\mathcal{O}(bpT)$ log-additions \cite{watanabe2017hybrid} while input-synchrony uses only $\mathcal{O}(bp)$ log-additions \cite{hannun2014first}.

In \Tref{tab:label_vs_time}, we perform an experimental validation of our theoretical understanding of the speed vs. accuracy trade-off between the two synchronous joint decoding variants.
To quantify speed, we compute the real-time factor (RTF) as the ratio of decoding time over the duration of input speech. 
To quantify accuracy beyond the BLEU metric, we compute the search error rate \cite{meister2020best} by counting the sequences for which the hypothesis has higher exact likelihood than the reference.
\textit{For the same beam size, output is slower but more accurate than input-synchronous.}
We conclude that input-synchrony may in fact be preferable in applications with latency constraints.

\paragraph{D2} \textit{Why did synchronous joint decodings outperform re-scoring decodings in \Tref{tab:out_domain}?}

There are a family of two-pass decoding algorithms \cite{watanabe2017hybrid, sainath2019two}, which also achieve joint decoding by first estimating the likelihoods of a subset of sequences $\mathcal{V}'$ with one module and then re-scoring the estimates with the other module.
In these approaches, the subset $\mathcal{V}'$ is determined asynchronously, meaning the joint likelihood is not considered until the re-scoring step; this delayed consideration of the joint likelihood is the main drawback compared to the synchronous approaches.
If the attentional decoder is used to determine $\mathcal{V}'$, then $\mathcal{V}'$ would suffer from exposure/label bias and the length problem (\Sref{sec:hybrid_asr}). 
On the other hand, if CTC is used to determine $\mathcal{V}'$, the lack of causal modeling in CTC leads to poor estimates of $\mathcal{V}'$ -- particularly for translation. 

\section{Limitations: Considering Latency}
\begin{table}[t]
  \centering
    \include{tables/limitations}
    \caption{Comparison of joint decoding and pure-attention RTFs across different beam sizes. $\%\Delta$ between the joint RTF and pure-attention RTF for the same beam size is shown, where positive $\%$'s indicate slow-downs and negative $\%$'s indicate speed-ups.}
    \label{tab:limitations}
\end{table}

There are several potential limitations pertaining to the increased computational overhead and latency of the joint modeling approach.
One concern is that joint decoding is much slower, but we found that input-synchronous joint decoding is actually faster than pure-attention decoding for smaller beam sizes (\Tref{tab:limitations}).
The other limitation is that our MT models upsample input representations in the early layers of the encoder, thereby increasing the computations in subsequent encoder layers and the decoder's cross-attention.
We can use our LegoNN-based encoder \cite{dalmia2022legonn} to adjust the up-sampling rate to a fractional value, minimizing the computations given dataset statistics.
Alternatively, we may avoid the need for up-sampling by applying a larger byte-pair encoding size \cite{kudo2018sentencepiece} to the target language compared to the source language. 

\section{Related Works}
\label{sec:prior_works}

The idea of using latent alignments to improve autoregressive translation has been explored previously by \citet{haviv2021can} who concluded that CTC alignments are not compatible with teacher forcing.
The key difference is that we train CTC and autoregressive models jointly while \citet{haviv2021can} sought to apply CTC to train autoregressive models, replacing cross-entropy entirely.
More recently in a concurrent work, \citet{zhang2022revisiting} have also shown the effectiveness of jointly training CTC and attention in the context of ST for un-written languages where not ASR transcriptions are available.
We believe that our contribution showing the effectiveness of also jointly decoding CTC and attention demonstrates an additional technique which can further improve their direction.
Our work also differs in that we seek to incorporate the ASR objective into ST via our hierarchical encoding scheme.
Other concurrent works, also integrated CTC and attention within blockwise streaming \cite{deng2022blockwise} and compositional multi-decoder \cite{yan-etal-2022-cmus} architectures for ST.
Our work supports their findings by addressing \textit{why} CTC is helping, and we provide a unified approach that generalizes to both MT and ST.

Prior works have also used the non-autoregressive property of CTC as means for speeding up autoregressive models during inference \cite{inaguma2021fast, gaido2021ctc}, but these works do not apply CTC to improve the translation quality of autoregressive models.
There is a significant line of prior work seeking to enhance attentional encoder-decoder MT models with lexical \cite{alkhouli2016alignment, song2020alignment}, syntactical \cite{wang2019source, bugliarello-okazaki-2020-enhancing, deguchi-etal-2021-synchronous}, and phrasal \cite{watanabe2021neural,zhang2021neural} alignment information derived from external discrete or latent knowledge bases.
These works do not investigate CTC in particular.

Finally, we believe that our approach can be generalized by applying other alignment approaches which are conceptually related to CTC, such as statisical \cite{jelinek1998statistical, koehn-etal-2003-statistical} and sequence aligning approaches \cite{graves2012sequence, ghazvininejad2020aligned, saharia-etal-2020-non}.

\section{Conclusion}
We propose to jointly train and decode CTC/attention models for MT and ST using 1) hierarchical encoding to resolve incompatibilities between CTC and the non-monotonic mappings in translation and 2) synchronous decoding to ease the exposure/label biases of autoregressive decoders with CTC's conditionally independent alignment information.
Our analyses reveal several reasons why even weak CTC models benefit autoregressive translation via joint modeling, suggesting that future explorations into jointly modeling attentional decoders with other latent alignment models may be fruitful.

\clearpage

\section*{Acknowledgements}
Brian Yan and Shinji Watanabe are supported by the Human Language Technology Center of Excellence.
This work used the Extreme Science and Engineering Discovery Environment (XSEDE) ~\cite{xsede}, which is supported by National Science Foundation grant number ACI-1548562; specifically, the Bridges system ~\cite{nystrom2015bridges}, as part of project cis210027p, which is supported by NSF award number ACI-1445606, at the Pittsburgh Supercomputing Center. 

\bibliography{anthology,custom}
\bibliographystyle{acl_natbib}

\clearpage

\appendix

\section{Supplementary Results}
\label{sec:v1}
\begin{table}[t]
  \centering
    \include{tables/v1}
    \caption{Comparison of our best MuST-C-v1 En-De Joint CTC/Attn model and our Pure-Attn baseline with prior works: $^1$\citet{inaguma2020espnet}, $^2$\citet{le2020dual}, $^3$\citet{du2022regularizing}, $^4$\citet{dalmia2021searchable}}
    \label{tab:v1}
\end{table}
\subsection{MuST-C-v1 Back-Compatibility}
See \Tref{tab:v1} for results compared to prior works.

\section{Reproducibility}
\label{sec:appendix}

\label{sec:repro}

\subsection{Dataset Descriptions}

\begin{table*}[t]
  \centering
    \include{tables/data}
    \caption{MT/ST/ASR dataset descriptions. Utterance counts are rounded to the nearest thousand. Language codes: De=German, En=English, Es=Spanish, Ja=Japanese, Fr=French, Pt=Portuguese, It=Italian, Ru=Russian, El=Greek}
    \label{tab:data}
\end{table*}

See \Tref{tab:data} for dataset descriptions.
Data preparation was done using ESPnet recipes:
IWSLT14\footnote{\url{https://github.com/espnet/espnet/tree/master/egs2/iwslt14/mt1}},
MuST-C\footnote{\url{https://github.com/espnet/espnet/tree/master/egs/must_c/st1}}, 
MTedX\footnote{\url{https://github.com/espnet/espnet/tree/master/egs/mtedx/st1}}.

\subsection{Model Architectures}

\begin{table*}[t]
  \centering
    \include{tables/models}
    \caption{MT/ST/ASR model descriptions. The best MT/ST Encoder layers settings were selected over a search space indicated by $\mathcal{S}$. Parameter counts are rounded to the nearest million. Note that the 12 layer pure-attn model outperformed the 18 layer version and that the 12 layer joint model still outperformed these baselines.}
    \label{tab:models}
\end{table*}

See \Tref{tab:models} for model architectures.

\begin{figure}[ht!]
    \centering
    \includegraphics[width=.9\linewidth]{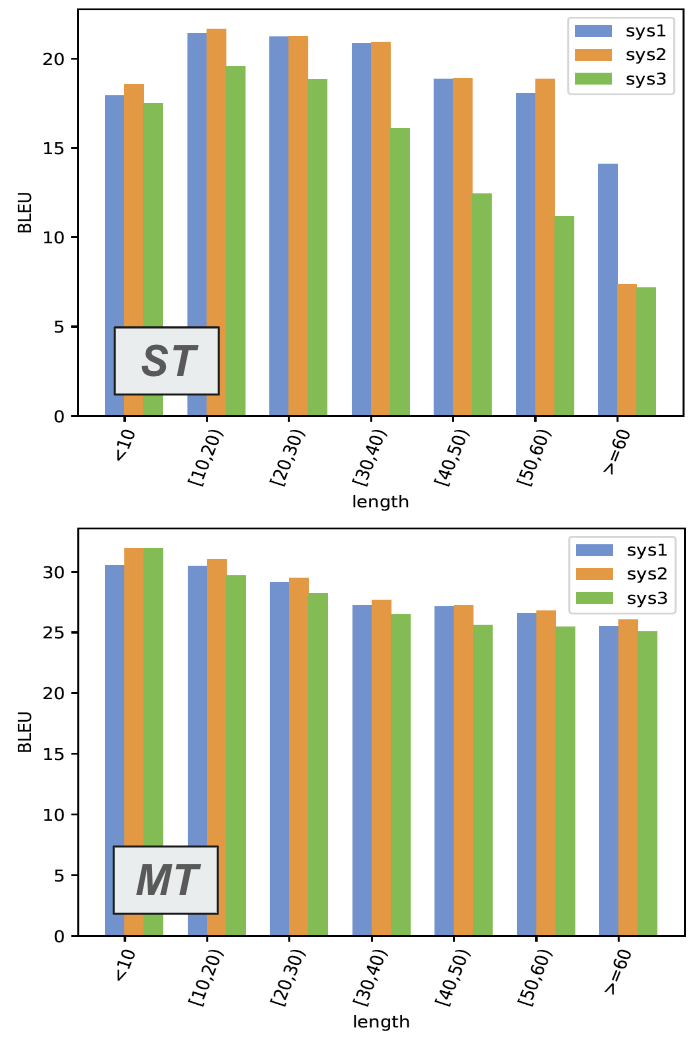}
    \caption{Compare-mt \cite{neubigcompare} output sentence length to BLEU for joint decoding vs pure-attention models. Model codes: sys1 = Joint Input-Sync, sys2 = Joint Output-Sync, sys3 = Pure-Attn }
    \label{fig:compare-mt}
\end{figure}

\subsection{Training/Decoding Hyperparameters}

\begin{table}[ht]
  \caption{Training Hyperparameters for MT Models.}
  \label{tab:hp-mt-tr}
  \centering
  \resizebox {\linewidth} {!} {
  \begin{tabular}{lr}
    \toprule
    Hyperparameter & Value \\
    \midrule
    Hidden Dropout & 0.3 \\
    Attention dropout & 0.3 \\
    Activation dropout & 0.3 \\
    LR schedule & inv. sqrt. \cite{vaswani2017attention}\\
    Max learning rate & best of [1e-3, 3e-3] \\
    Warmup steps & 10000 \\
    Number of steps & 200 epoch \\
    Adam eps  & 1e-9 \\
    Adam betas  & (0.9, 0.98)\\
    Weight decay & 1e-4\\
    $\lambda1$, $\lambda2$ & (1, 2)\\
    \bottomrule
  \end{tabular}
  }
\end{table}

\begin{table}[ht]
  \caption{Training Hyperparameters for ST Models.}
  \label{tab:hp-st-tr}
  \centering
  \resizebox {\linewidth} {!} {
  \begin{tabular}{lr}
    \toprule
    Hyperparameter & Value \\
    \midrule
    Hidden Dropout & 0.1 \\
    Attention dropout & 0.1 \\
    Activation dropout & 0.1 \\
    LR schedule & inv. sqrt. \cite{vaswani2017attention}\\
    Max learning rate & 0.002 \\
    Warmup steps & 25000 \\
    Number of steps & 40 epoch \\
    Adam eps  & 1e-9 \\
    Adam betas  & (0.9, 0.98)\\
    Weight decay & 0.0001\\
    $\lambda1$, $\lambda2$ & (2, 5)\\
    \bottomrule
  \end{tabular}
  }
\end{table}

\begin{table}[ht]
  \caption{Decoding Search Space MT Models.}
  \label{tab:hp-mt-de}
  \centering
  \resizebox {\linewidth} {!} {
  \begin{tabular}{llr}
    \toprule
Decoding Type  & Hyperparameter & Value \\
    \midrule
\multirow{2}{*}{Pure Attn} &    Max Length Ratio &  [1, 1.2, 1.4, 1.6, 1.8, 2, 2.5, 3] \\
    & Penalty & [0, 0.2, 0.4, 0.6, 0.8, 1.0]\\
    & Beam Size & 5 \\
    \midrule
 \multirow{3}{*}{Joint O-Sync}   & Max Length Ratio & 1 \\
    & Penalty & [0, 0.2, 0.4, 0.6, 0.8, 1.0]\\
    & CTC Weight & 0.3 \\
    & Beam Size & 5 \\
    \midrule
 \multirow{4}{*}{Joint I-Sync}  & Max Length Ratio & 1 \\
    & Penalty & [0, 0.2, 0.4, 0.6, 0.8, 1.0]\\
    & Blank Penalty & [0.5, 0.75, 1.0]\\
    & CTC Weight & [0.3, 0.5]\\
    & Beam Size & [10, 30]\\
    \bottomrule
  \end{tabular}
  }
\end{table}

\begin{table}[ht]
  \caption{Decoding Search Space ST Models.}
  \label{tab:hp-st-de}
  \centering
  \resizebox {\linewidth} {!} {
  \begin{tabular}{llr}
    \toprule
Decoding Type  & Hyperparameter & Value \\
    \midrule
\multirow{2}{*}{Pure Attn} & Max Length Ratio & 1 \\
    & Penalty & [0,0.2,0.4,0.6,0.8,1.0]\\
    & Beam Size & [10, 30, 50]\\
    \midrule
 \multirow{3}{*}{Joint O-Sync} & Max Length Ratio & 1 \\
    & Penalty & [0,0.2,0.4,0.6,0.8,1.0]\\
    & CTC Weight & [0.3, 0.5] \\
    & Beam Size & [10, 30, 50]\\
    \midrule
 \multirow{4}{*}{Joint I-Sync} & Max Length Ratio & 1 \\
    & Penalty & [0,0.2,0.4,0.6,0.8,1.0]\\
    & Blank Penalty & 1\\
    & CTC Weight & [0.3, 0.5]\\
    & Beam Size & [10, 30, 50]\\
    \bottomrule
  \end{tabular}
  }
\end{table}

See \Tref{tab:hp-mt-tr} for MT training, \Tref{tab:hp-st-tr} for ST training, \Tref{tab:hp-mt-de} for MT decoding, and \Tref{tab:hp-st-de} for ST decoding hyperpameters.

\subsection{Metrics}

Sacrebleu signature for all non-Japanese: 

$\texttt{BLEU+case.mixed+numrefs.1}$
$\texttt{+smooth.exp+tok.13a+version.1.5.1}$
\\
\\
\noindent Sacrebleu signature for Japanese:

$\texttt{BLEU+case.mixed+lang.en-ja+numrefs.1}$
$\texttt{+smooth.exp+tok.ja-mecab-0.996-IPA}$
$\texttt{+version.1.5.1}$
\\
\\
\noindent For tokenized BLEU in the IWSLT MT datasets we used $\texttt{mutibleu.perl}$ \cite{multibleu}

\subsection{Computing}

ST models were trained on 2 x V100 for 2 days. MT models were trained on 1 x A6000 for 1 day.

\subsection{Valid Set performances}
\Tref{tab:main_valid} presents the validation performances for our ST and MT models.
\begin{table*}[t]
  \centering
    \include{tables/main_valid}
    \caption{Valid set performances, as measured by BLEU ($\uparrow$), of our proposed joint CTC/Attention models compared to pure-attention baselines. Joint CTC/Attention models are always jointly trained, but can be either jointly decoded using input/output synchronony or decoded using only their CTC or attention branches.}
    \label{tab:main_valid}
\end{table*}

\section{Additional Analyses}

\subsection{$\operatorname{compare\_mt.py}$ Analysis}

As shown in \Fref{fig:compare-mt}, both joint synchronous decodings are more robust than pure-attention for long output lengths. Input-synchrony appears most robust in generation of very long outputs for ST.

\subsection{View of Regularized Attention}

\begin{figure*}
    \centering
    \includegraphics[width=\linewidth]{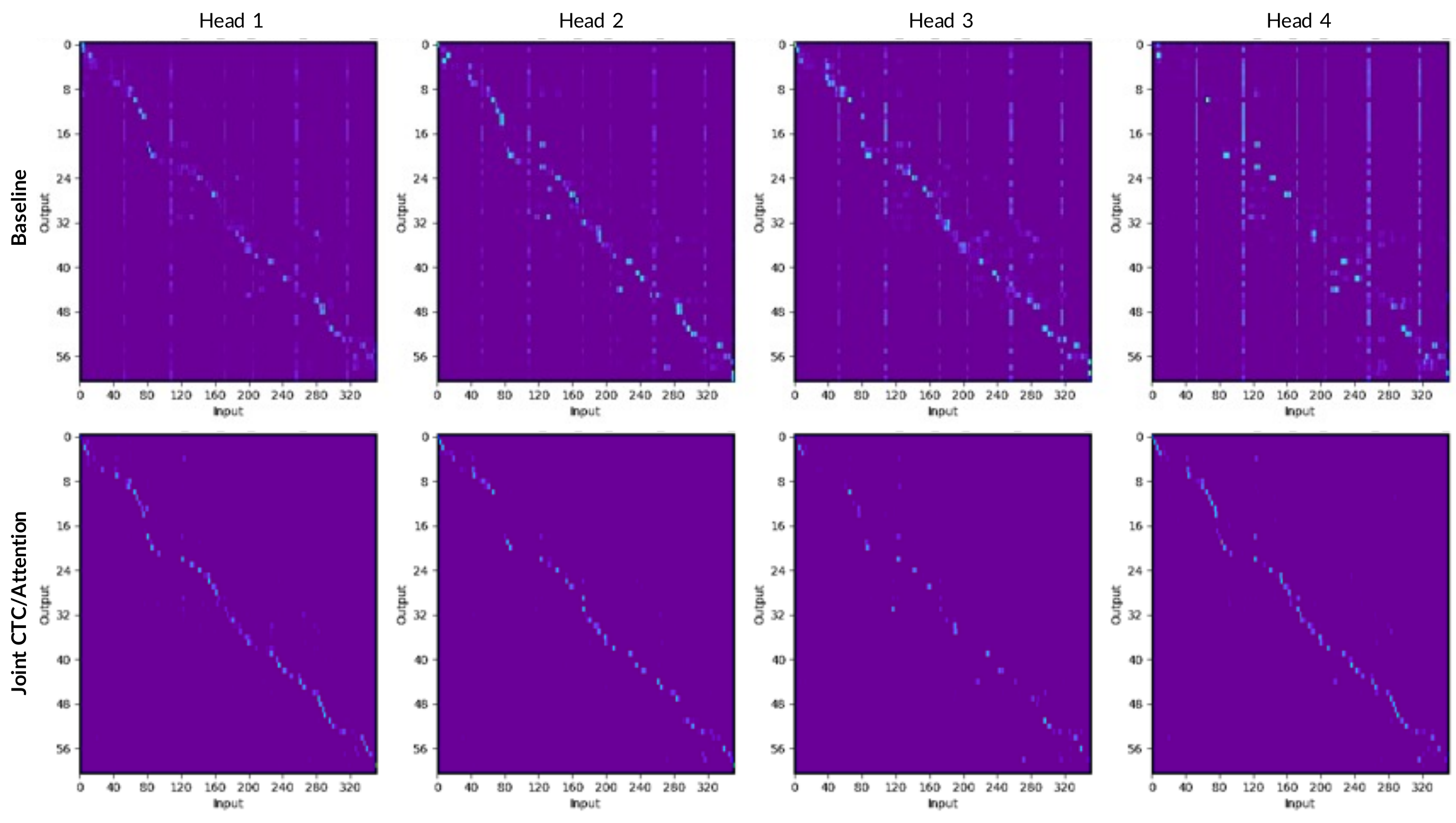}
    \caption{Visualization of source attention patterns produced by pure-attention baseline (top) vs. joint CTC/attention (bottom) ST models. Qualitative example extracted from the final decoder layer. Irregular patterns are observable in the pure-attention plots, but not in the joint CTC/attention plots.}
    \label{fig:qual_attn}
\end{figure*}

See \Fref{fig:qual_attn} for a qualitative example of monotonic source attention patterns (supplementary to the quantitative monotonicity in \Tref{fig:mono}).

\end{document}

%% file: tables/conjectures.tex
\resizebox {\linewidth} {!} {
\begin{tabular}{p{0.4\linewidth} | p{0.4\linewidth} | p{0.5\linewidth} | p{0.05\linewidth} | p{0.07\linewidth}}
\toprule
\textsc{CTC} & \textsc{Attention} & \textsc{Joint CTC/Attention} & \textsc{ASR} & \textsc{MT/ST} \\
\midrule
$P_\text{CTC}(Y|X) \overset{\Delta}{=} \displaystyle\sum_{Z\in \mathcal{Z}} \prod_{t=1}^{T} P(z_t | X, \cancel{z_{1:t-1}}) $ & $P_{\text{Attn}}(Y|X) \overset{\Delta}{=} \prod_{l=1}^{L} P(y_l|y_{1:l-1}, X)$ & $P_{\text{Joint}}(Y|X) \overset{\Delta}{=}P_\text{CTC}(Y|X)^{\lambda} \times P_\text{Attn}(Y|X)^{1-\lambda}$ & \cmark & \cmark\\
\midrule
\textcolor{blue}{\textbf{Hard Alignment}} \textcolor{white}{.....................................} Criterion only allows monotonic alignments of inputs to outputs & \textcolor[HTML]{b2182b}{\textbf{Soft Alignment}} \textcolor{white}{........................................} Flexible attention-based input-to-output mappings may overfit to irregular patterns & \textbf{During Training:} \textcolor{blue}{Hard alignment objective produces stable encoder representations} \textcolor[HTML]{b2182b}{allowing the decoder to more rapidly learn soft alignment patterns} & \cmark & \textbf{L1}\textcolor{white}{..} See \Sref{sec:ctc_limitations} \\
\midrule
\textcolor{blue}{\textbf{Conditional Independence}} \textcolor{white}{....................} Assumes that there are no dependencies between each output unit given the input & \textcolor[HTML]{b2182b}{\textbf{Conditional Dependence}} \textcolor{white}{.......................} Locally normalized models with output dependency exhibit label/exposure biases & \textbf{During Decoding:} \textcolor{blue}{Use of conditionally independent likelihoods in joint scoring} \textcolor[HTML]{b2182b}{eases the exposure/label biases from conditionally dependent likelihoods}  & \cmark & \textbf{L2}\textcolor{white}{..} See \Sref{sec:ctc_limitations} \\
\midrule
\textcolor{blue}{\textbf{Input-Synchronous Emission}} \textcolor{white}{...............} Each input representation emits exactly one blank or non-blank output token & \textcolor[HTML]{b2182b}{\textbf{Autoregressive Generation}} \textcolor{white}{....................} Need to detect end-points and compare hypotheses of different length in beam search & \textbf{During Decoding:} \textcolor{blue}{Input-synchronous emission determines output length based on input length} \textcolor[HTML]{b2182b}{counteracting the autoregressive end-detection problem}  & \cmark & \textbf{L3}\textcolor{white}{..} See \Sref{sec:ctc_limitations} \\
\bottomrule
\end{tabular}
}

%% file: tables/main.tex
\resizebox {\linewidth} {!} {
\begin{tabular}{l|ccc|ccc|ccc}
\toprule
& \multicolumn{3}{c}{\textsc{Model Type}} & \multicolumn{3}{|c|}{\textsc{MT}} & \multicolumn{3}{c}{\textsc{ST}}  \\
\cmidrule(r){2-4}\cmidrule(r){5-7}\cmidrule(r){8-10}
& Joint & Joint & Decoding & IWSLT14 & IWSLT14 & MTedX & MuST-C-v2 & MuST-C-v2 & MTedX \\
\textsc{Model Name} & Train? & Decode? & Method & De-En & Es-En & All-En & En-De & En-Ja & All-En \\
\midrule
Pure-Attn (Prior)  & \xmark & \xmark & Attn Only & (32.15)$^\dag$ & (38.95)$^\dag$ & -$^\diamondsuit$ & 25.8$^\ddag$ & 12.4$^\ddag$ & -$^\diamondsuit$  \\
Pure-Attn (Ours) & \xmark & \xmark & Attn Only & 32.8 (33.73) & 39.0 (39.86) & 25.6 & 27.8 & 14.3 & 22.7 \\
\midrule
Joint CTC/Attn & \cmark & \xmark & CTC Only & 27.3 & 33.8 & 22.4 & 24.4 & 10.2 & 21.4 \\
Joint CTC/Attn & \cmark & \xmark & Attn Only & 33.6 & 39.5 & 28.0 & 28.3 & 14.2 & 23.7 \\
\midrule
Joint CTC/Attn & \cmark & \cmark & Joint I-Sync & 33.7 & 39.7 & 27.8 & \textbf{29.2} & 15.1 & \textbf{25.1} \\
Joint CTC/Attn & \cmark & \cmark & Joint O-Sync & \textbf{34.1} & \textbf{39.9} & \textbf{28.1} & \textbf{29.2} & \textbf{15.3} & \textbf{25.1}\\
\bottomrule
\end{tabular}
}

%% file: tables/ablation.tex
\resizebox {0.9\linewidth} {!} {
\begin{tabular}{cc|c|c}
\toprule
& & \textsc{MT (De-En)} & \textsc{ST (En-De)} \\
\cmidrule(r){3-3}\cmidrule(r){4-4}
\textsc{SrcCTC} & \textsc{TgtCTC} & IWSLT14 & MuST-C-v2 \\
\midrule
\xmark & \xmark & 32.1 & 27.7 \\
\cmark & \xmark & 34.1 & 27.8 \\
\xmark & \cmark & 33.3 & 28.1 \\
\cmark & \cmark & \textbf{34.8} & \textbf{28.3} \\
\bottomrule
\end{tabular}
}

%% file: tables/out_domain.tex
\resizebox {\linewidth} {!} {
\begin{tabular}{l|c|cc|cc}
\toprule
& \textsc{Type} &\multicolumn{2}{c|}{\textsc{MT (De-En)}} & \multicolumn{2}{c}{\textsc{ST (En-De)}} \\
\cmidrule(r){2-2}\cmidrule(r){3-4}\cmidrule(r){5-6}
& Decoding & IWSLT14 & EuroParl & MuST-C-v2 & EuroParl\\
\textsc{Model} & Method & (In-D) & (Out-D) & (In-D) & (Out-D) \\
\midrule
Pure-Attn & Attn Only & 32.8 & 15.8 & 27.8 & 20.5 \\
\midrule
Joint C/A & Attn Only & 33.6 & 17.1 & 28.3 & 21.0 \\
&+CTC Rescore & 33.6 & 17.1 & 28.3 & 21.0 \\
Joint C/A & Joint O-Sync & \textbf{34.1} & \textbf{17.6} & \textbf{29.2} & \textbf{21.7} \\
\midrule
Joint C/A & CTC Only & 27.3 & 13.1 & 24.4 & 16.5 \\
&+Attn Rescore & 29.5 & 13.9 & 26.2 & 17.8 \\
Joint C/A & Joint I-Sync & \textbf{33.7} & \textbf{17.4} & \textbf{29.2} & \textbf{21.1} \\
\bottomrule
\end{tabular}
}

%% file: tables/label_vs_time.tex
\resizebox {.95\linewidth} {!} {
\begin{tabular}{lc|cc|c}
\toprule
\multicolumn{2}{c|}{Decoding Type} & \multicolumn{2}{c|}{Accuracy} & Speed \\
\cmidrule{1-2}\cmidrule{3-4}\cmidrule{5-5}
Method & Beam Size & BLEU & Search Error & RTF  \\
\midrule
Joint O-Sync & 5 & 29.1 & 0.73\% & 0.9 \\
Joint O-Sync & 10 & 29.2 & 0.44\% & 1.7 \\
Joint O-Sync & 50 & 29.0 & 0.36\% & 9.0 \\
\midrule
Joint I-Sync & 5 & 28.1 & 1.02\% & 0.4 \\
Joint I-Sync & 10 & 28.6 & 1.09\% & 0.9 \\
Joint I-Sync & 50 & 29.0 & 0.87\% & 6.4 \\
\bottomrule
\end{tabular}
}

%% file: tables/limitations.tex
\resizebox {.95\linewidth} {!} {
\begin{tabular}{lc|cc}
\toprule
\multicolumn{2}{c|}{Decoding Type} & \multicolumn{2}{c}{Speed} \\
\cmidrule{1-2}\cmidrule{3-4}
Method & Beam Size & RTF & $\%\Delta$ \\
\midrule
Pure-Attn O-Sync & 5 & 0.9 & -\\
Pure-Attn O-Sync & 10 & 1.2 & - \\
Pure-Attn O-Sync & 50 & 3.5 & -\\
\midrule
Joint CTC/Attn O-Sync & 5 & 0.9 & +0$\%$ \\
Joint CTC/Attn O-Sync & 10 & 1.7 & +42$\%$ \\
Joint CTC/Attn O-Sync & 50 & 9.0 & +157$\%$ \\
\midrule
Joint CTC/Attn I-Sync & 5 & 0.4 & -56$\%$\\
Joint CTC/Attn I-Sync & 10 & 0.9 & -25$\%$\\
Joint CTC/Attn I-Sync & 50 & 6.4 & +85$\%$ \\
\bottomrule
\end{tabular}
}

%% file: tables/v1.tex
\resizebox {.95\linewidth} {!} {
\begin{tabular}{l|c}
\toprule
& MuST-C-v1 \\
\textsc{Model Name} & En-De \\
\midrule
ESPnet-ST$^1$ & 22.9  \\
Dual-Decoder$^2$ & 23.6 \\
E2E-ST-TDA$^3$ & 25.4 \\
Multi-Decoder$^4$ & 26.4 \\
Pure-Attn (ours) & 27.1 \\
\midrule
Joint CTC/Attn w/ Joint O-Sync & \textbf{28.2} \\
\bottomrule
\end{tabular}
}

%% file: tables/data.tex
\resizebox {\linewidth} {!} {
\begin{tabular}{lcccccc}
\toprule
Dataset & Task & Source Lang(s) & Target Lang(s) & Domain & \# Train/Valid/Test Utts & \# Speech Train Hours \\
\midrule
IWSLT17 \cite{cettolo2012wit3} & MT & De & En & TED Talk & 160k/7k/7k & - \\
IWSLT17 \cite{cettolo2012wit3} & MT & De & Es & TED Talk & 160k/7k/7k & - \\
\midrule
MuST-C-v2 \cite{di-gangi-etal-2019-must} & ASR/ST & En & De & TED Talk & 250k/1k/3k & 450h \\
MuST-C-v2 \cite{di-gangi-etal-2019-must} & ASR/ST & En & Ja & TED Talk & 330k/1k/3k & 540h \\
\midrule
MTedX \cite{salesky2021multilingual} & MT & Es, Fr, Pt, It, Ru, El & En & TED Talk & 130k/6k/6k & -  \\
MTedX \cite{salesky2021multilingual} & ASR & Es, Fr, Pt, It, Ru, El & En & TED Talk & 400k/6k/6k & 730h \\
MTedX \cite{salesky2021multilingual} & ST & Es, Fr, Pt, It, Ru, El & En & TED Talk & 130k/6k/6k & 250h \\
\midrule
EuroParl \cite{iranzo2020europarl} & MT & De & En & Parliament Speech & - /-/2k & - \\
EuroParl \cite{iranzo2020europarl} & ST & En & De & Parliament Speech & -/-/1k & -\\
\bottomrule
\end{tabular}
}

%% file: tables/models.tex
\resizebox {\linewidth} {!} {
\begin{tabular}{lccccccccc}
\toprule
Model & Task & \# Encoder Layers [$\mathcal{S}$] & \# Decoder Layers & $\operatorname{SrcCTC}$ Layer & Up/Down-Sample & Pre-Train Init & Src BPE Size & Tgt BPE Size & \# Params \\
\midrule
Pure-Attn & MT & 12 [6,12,18] & 6 & - & - & - & \multicolumn{2}{c}{10k (joint)} & 54M \\
Joint CTC/Attn & MT & 18 [6,12,18] & 6 & 6 & 3x &  - & \multicolumn{2}{c}{10k (joint)} & 95M \\
\midrule
Pure-Attn & ST & 18 [12, 18] & 6 & - & 1/4x & Enc lyr 1-12 from ASR & 4k & 4k & 74M \\
Joint CTC/Attn & ST & 18 [12, 18] & 6 & 12 & 1/4x & Enc lyr 1-12 from ASR & 4k & 4k & 72M \\
\midrule
Pure-Attn & ASR & 12 & 6 & - & - & - & 4k & 4k & 46M \\
\bottomrule
\end{tabular}
}

%% file: tables/main_valid.tex
\resizebox {.95\linewidth} {!} {
\begin{tabular}{l|ccc|cc|cc}
\toprule
& \multicolumn{3}{c}{Model Type} & \multicolumn{2}{|c|}{MT} & \multicolumn{2}{c}{ST}  \\
\cmidrule(r){2-4}\cmidrule(r){5-6}\cmidrule(r){7-8}
& Joint & Joint & Decoding & IWSLT14 & IWSLT14 & MuST-C-v2 & MuST-C-v2 \\
Model Name & Train? & Decode? & Method & De-En & Es-En & En-De & En-Ja \\
\midrule
Pure-Attn (Ours) & \xmark & \xmark & Attn O-sync & 34.1 & 41.2 & 28.5 & 11.3 \\
\midrule
Joint CTC/Attn & \cmark & \cmark & Joint I-sync & 34.6 & 42.0 & 29.0 & 12.4 \\
Joint CTC/Attn & \cmark & \cmark & Joint O-sync & 35.0 & 42.3 &  29.2 & 12.4 \\
\bottomrule
\end{tabular}
}